\pdfoutput=1

\documentclass[11pt]{article}

\usepackage[preprint]{acl}

\usepackage{times}
\usepackage{latexsym}

\usepackage[T1]{fontenc}

\usepackage[utf8]{inputenc}

\usepackage{microtype}

\usepackage{inconsolata}

\usepackage{graphicx}
\usepackage{makecell}
\usepackage{enumitem}
\usepackage{multirow}
\usepackage{epsfig}
\usepackage{epstopdf}
\usepackage{booktabs}
\usepackage{tabularx}
\usepackage{tcolorbox}
\usepackage{colortbl}
\usepackage{subfigure}
\usepackage[utf8]{inputenc}
\usepackage{subcaption}
\newcommand{\samethanks}[1][\value{footnote}]{\footnotemark[#1]}
\newcommand{\equalcontribution}{\thanks{Equal contribution.}}

\title{HARE: HumAn pRiors, a key to small language model Efficiency}

\author{
 \textbf{Lingyun Zhang}\equalcontribution,
 \textbf{Bin jin}\samethanks,
 \textbf{Gaojian Ge},
 \textbf{Lunhui Liu},
 \\
 \textbf{Xuewen Shen},
 \textbf{Mingyong Wu},
 \textbf{Houqian Zhang},
 \textbf{Yongneng Jiang},
 \textbf{Shiqi Chen},
 \textbf{Shi Pu}\samethanks\thanks{Corresponding author.}
\\
 LiteAI China Telecom Guizhou Branch
\\
 \small{
    \{zhangly41, jinb1, gegj, liulh5, shenxw, wumy16, zhanghq20, jiangyn4, chensq27, pus1\}@chinatelecom.cn
 }
}

\begin{document}
\maketitle
\begin{abstract}
Human priors play a crucial role in efficiently utilizing data in deep learning.
However, with the development of large language models (LLMs), there is an increasing emphasis on scaling both model size and data volume, which often diminishes the importance of human priors in data construction. 
Influenced by these trends, existing Small Language Models (SLMs) mainly rely on web-scraped large-scale training data, neglecting the proper incorporation of human priors. 
This oversight limits the training efficiency of language models in resource-constrained settings.
In this paper, we propose a principle to leverage human priors for data construction.
This principle emphasizes achieving high-performance SLMs by training on a concise dataset that accommodates both semantic diversity and data quality consistency, while avoiding benchmark data leakage.
Following this principle, we train an SLM named HARE-1.1B \footnote{\url{https://github.com/LiteAI-Team/HARE}}. 
Extensive experiments on large-scale benchmark datasets demonstrate that HARE-1.1B performs favorably against state-of-the-art SLMs, validating the effectiveness of the proposed principle. 
Additionally, this provides new insights into efficient language model training in resource-constrained environments from the view of human priors.
\end{abstract}

\section{Introduction}
Small language models (SLMs) have recently garnered significant attention for their computational efficiency and real-time responsiveness \citep{mehta2024openelm, zhang2024tinyllama, li2023textbooks, singer2024h2odanube18b}. 
However, influenced by the scaling law of large language models (LLMs), existing SLMs \citep{mehta2024openelm, zhang2024tinyllama} rely heavily on web-scraped large-scale data.
This reliance limits their efficient training in resource-constrained settings due to inconsistent quality and a lack of semantic diversity in the data. 
With limited parameters, SLMs face significant training challenges compared to LLMs \citep{radford2019language}.
To mitigate these issues, \citep{li2023textbooks} and \citep{benallal2024cosmopedia} synthesize textbook-quality training data using high-performance LLMs, achieving competitive performance with less training data. 
\begin{figure}[!tbp]
  \centering
  \includegraphics[width=\columnwidth]{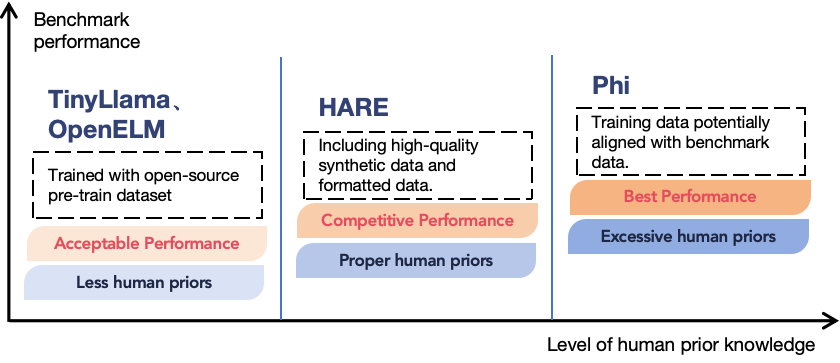}
  \caption{
    \label{fig:model_class}
    Illustration showing the performances of representative SLMs \citep{zhang2024tinyllama, mehta2024openelm, li2023textbooks} on the benchmark datasets in the open LLM leaderboard \citep{open-llm-leaderboard} at different levels of human prior knowledge. As the integration of human priors deepens, models demonstrate improved performance, but with an increased risk of data leakage.
  }
\vspace{-0.5cm}
\end{figure}
The success of \citep{li2023textbooks} and \citep{benallal2024cosmopedia} can be attributed to incorporating human priors in data construction. 
These methods cluster large-scale web-scraped data into multiple topics and use diverse prompts along with topic-specific data as seed data for data synthesis. 
This process enhances semantic diversity, reflecting the human prior that diversity is crucial for improving model generalization.
Additionally, using LLMs to filter and govern web-scraped data ensures consistent data quality. 
This aligns with human priors on the need for data consistency to avoid the negative impacts of noise and errors on model training.
This motivation to incorporate human priors parallels the approach of effective supervised learning models \citep{xia2021using,von2021informed}, where increasing sample diversity and applying consistent labels enhance the quality of the training data. 
By embedding these human priors into the training data, models trained under better conditions, thus enhancing performance and generalization.
\begin{figure*}[!thbp]
  \vspace{-0.45cm}
  \centering
  \includegraphics[width=0.9\linewidth]{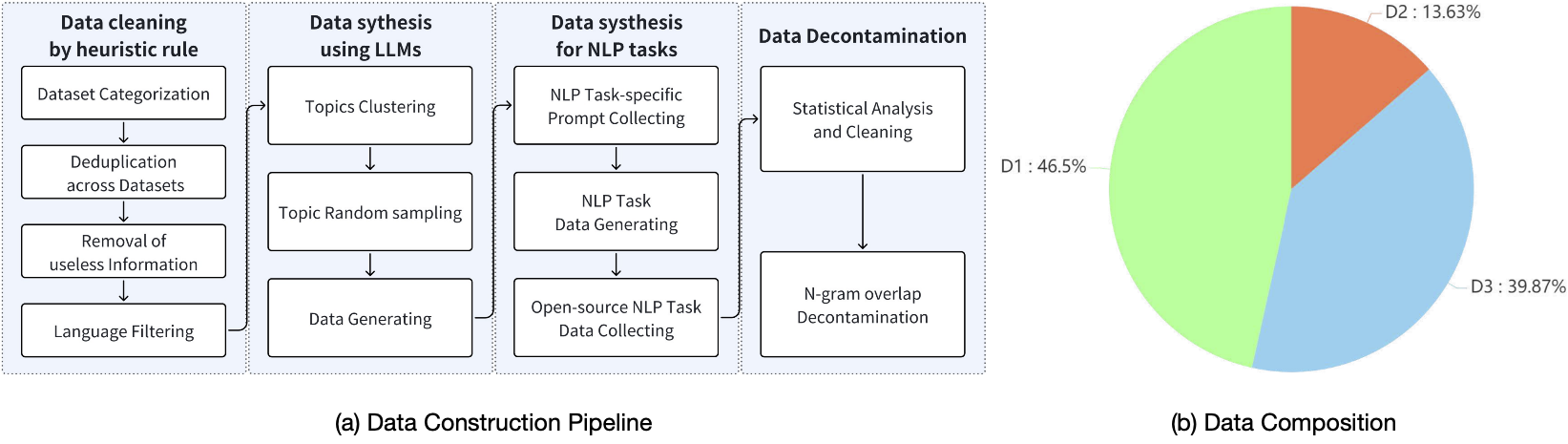}
  \caption{
  \label{fig:pipeline}\label{fig:dataset}
    Data Construction Pipeline and Composition.
    $D_1$: high-quality categorized open-source cleaned data.
    $D_2$: semantic enriched synthetic data.
    $D_3$: mixture of synthetic and open-source NLP task data.
  }
\vspace{-0.5cm}
\end{figure*}
The incorporation of human priors helps \citep{li2023textbooks} achieve outstanding performance on benchmark datasets.
However, we observe that some recent SLMs \citep{bai2023qwen,li2023textbooks,bellagente2024stable} may inject excessive human priors, risking benchmark data leakage as shown in Figure \ref{fig:model_class} and Table \ref{tab:data leak}. 
Based on our analysis of human priors in existing works, we propose a data construction principle: ensuring both semantic diversity and data quality consistency while avoiding benchmark data leakage.
We noted that \citep{benallal2024cosmopedia} is the most similar to our proposed principle. 
However, this approach does not achieve satisfactory performance on benchmark evaluations.
This is because, while injecting semantic diversity, it does not incorporate Natural Language Processing (NLP) task semantics in natural language form, resulting in a lack of NLP-task solving capability on benchmarks. 
\citep{radford2019language} and \citep{NEURIPS2020_1457c0d6} demonstrate that LLMs can learn zero-shot capabilities for NLP tasks from vast amounts of data. For SLMs to acquire the same capability from limited data, they must rely on the injection of human priors.

Therefore, guided by this principle, we propose our data construction method: 
(1) Extract high-quality, clearly categorized data from large-scale web-scraped datasets to ensure semantic diversity and maintain consistently high data quality.
(2) Cluster large-scale web-scraped data into various topics and use topic-specific data with diverse prompts to generate synthetic data using high-performance LLMs, which enhance semantic diversity while ensuring consistent data quality.
(3) Construct large-scale NLP-task data in natural language form to enhance semantic diversity and improve NLP-task solving capabilities. 
(4) Implement strict decontamination procedures to avoid benchmark data leakage.
Using this method, we constructed a training dataset and trained an SLM named HARE-1.1B. 
This model performs favorably against existing state-of-the-art (SOTA) SLMs on large-scale benchmark datasets, validating our proposed principle and data construction method. 
We hope this work will inspire further exploration in incorporating human priors through the design of network architecture, loss functions, and regularization to enhance SLM training efficiency, not just through efficient data utilization.

\section{Data Construction}
\label{Data Construction}
In this section, we detail our data construction pipeline, as illustrated in Figure~\ref{fig:pipeline}(a).

\noindent\textbf{Data cleaning by heuristic rule.}
\label{sec:2.1}
We implement a series of heuristic rules to clean collected open-source pre-training corpora, including RedPajama \citep{together2023redpajama}, Pile \citep{gao2020pile}, OpenWebMath \citep{paster2023openwebmath}, and others. Appendix~\ref{Collected open-source data} provides a comprehensive list of our collected data. We categorize these datasets according to their sources and adjust proper sampling weights to enhance semantic diversity. The heuristic rules we apply include deduplication across datasets, removal of personal privacy information and web links, elimination of semantically incomplete samples, and exclusion of non-English samples. These processes ensure that the cleaned data consistently maintain high quality. We designate the cleaned data as $D_1$.

\noindent\textbf{Data synthesis using LLMs.}
Even after heuristic rule-based cleaning, unresolved semantic ambiguities, such as missing text paragraphs or formatting errors, may still exist. Therefore, we utilize the Mixtral-8$\times$7B \citep{jiang2024mixtral} model for data synthesis based on a subset extracted from $D_1$. Appendix~\ref{Data synthesis using LLMs} provides details of this subset.
First, we cluster the subset data into various topics and sample seed data from these topics. We then input diverse prompts coupled with the seed data into the Mixtral-8$\times$7B model for data synthesis. Appendix~\ref{Data synthesis using LLMs} presents the details of the data synthesis process.
This approach significantly enhances semantic diversity through diverse topics and prompts, while using the same LLM for data synthesis ensures consistent data quality. We set the synthesized data as $D_2$.

\noindent\textbf{Data synthesis for NLP tasks.}
To enhance NLP-task solving capabilities, we construct a substantial amount of NLP-task data in natural language form. Specifically, we create numerous prompts and use a subset of $D_2$ as seed data to guide the synthesis of diverse NLP-task data using Qwen1.5-32B \cite{bai2023qwen}. Additionally, we collect various open-source NLP-task datasets to further expand this dataset. Appendix~\ref{Mixture of Synthetic and Open-source NLP Task Data} provides details of the data synthesis process. We refer to this dataset for NLP tasks as $D_3$.

\noindent\textbf{Data decontamination.} 
To ensure that our generated data poses no risk of benchmark data leakage, we establish a rigorous data decontamination process. First, we conduct statistical analyses to remove samples that did not meet our standards, such as those with excessively short or long context lengths. Subsequently, we calculate the n-gram overlap with benchmark data to eliminate potentially duplicated samples. Appendix~\ref{Data Decontamination Process} details the decontamination process.

Our final training dataset consists of $D_1$, $D_2$, and $D_3$. Figure \ref{fig:dataset}(b) shows the sample weights of the three parts.

\section{Training}
\label{Pre-train}
We use the Mistral \citep{jiang2023mistral} architecture and reduce the model parameters to approximately 1.1B. The modified model consists of 22 hidden layers, 32 attention heads, and 8 key-value heads, with a hidden size of 2048 and a vocabulary size of 32,000. We name the pre-trained model HARE-1.1B.
HARE-1.1B is trained on 16 Nvidia-H800 GPUs using DeepSpeed \citep{rasley2020deepspeed} and Flash-Attention \citep{dao2022flashattention}. The maximum sequence length is set to 2048, with a batch size of 2 million tokens and a maximum learning rate of 3e-4. The training process is divided into two stages based on different data sources. In the first stage, we train our model on the $D_1$ dataset, processing 52 billion tokens. In the second stage, HARE-1.1B is trained on our final training dataset, which includes $D_1$, $D_2$, and $D_3$. The entire training process spans 30 days, during which 600 billion tokens are processed. Note that the number of processed tokens is far smaller than that of \citep{zhang2024tinyllama} (3T tokens), \citep{mehta2024openelm} (1.8T tokens), and \citep{singer2024h2odanube18b} (2T tokens).
Additionally, we have fine-tuned the trained 1.1B model for chat and android api calling applications, details can be found in the Appendix~\ref{Supervised Fine-tuning}.

\section{Experiments}
\label{Result}
In this section, we conduct ablation studies and comparisons with SOTA SLMs, including Phi1.5 \citep{li2023textbooks}, Qwen1.5 \citep{bai2023qwen}, Stablelm2 \citep{bellagente2024stable}, H2o-danube \citep{singer2024h2odanube18b}, openELM \citep{mehta2024openelm}, Csg-wukong \citep{opencsg2024wukong}, Cosmo \citep{benallal2024cosmopedia}, TinyLlama \citep{zhang2024tinyllama}, and Gpt2xl \citep{radford2019language}. These studies are performed on the Open LLM Leaderboard \citep{open-llm-leaderboard} using the MMLU \citep{hendrycks2021measuring}, ARC-C \citep{clark2018think}, TruthfulQA \citep{lin2022truthfulqa}, Winogrande \citep{DBLP:journals/corr/abs-1907-10641}, Hellaswag \citep{zellers2019hellaswag}, and Gsm8k \citep{DBLP:journals/corr/abs-2110-14168} benchmark datasets. We follow the standard benchmark protocols on the Open LLM Leaderboard.
\begin{figure}[htb]
  \centering
  \includegraphics[width=0.9\columnwidth,height=0.7\linewidth]{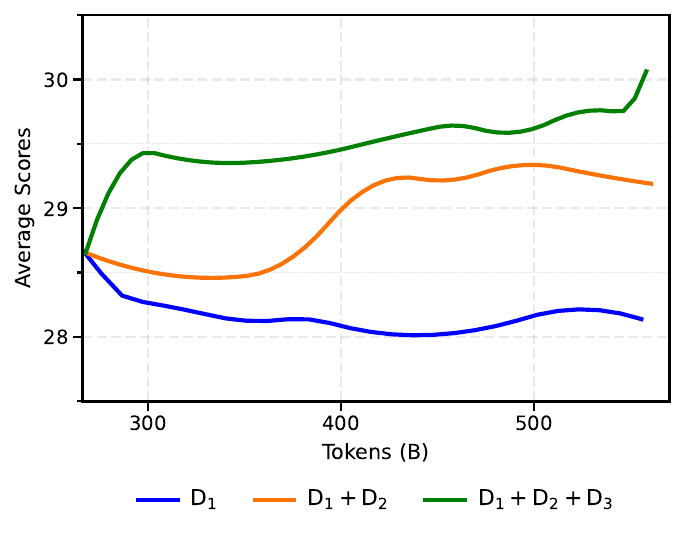}
  \caption{\label{fig:ablation_test}
    Ablation studies showing the average score curves of 0.25B models on the Open LLM Leaderboard as the number of training tokens increases.}
  \vspace{-0.6cm}
\end{figure}
\begin{table*}[t]
  \vspace{-0.45cm}
  \centering
  \small
  \begin{tabular}{c|cccccccc}
    \toprule[2pt]
    \textbf{Model} & \textbf{Size} & \textbf{Average} & \textbf{MMLU} & \textbf{ARC-C} & \textbf{TruthfulQA} & \textbf{Winogrande} & \textbf{Hellaswag} & \textbf{Gsm8k} \\
    \midrule[1pt]
    Phi1.5 & 1.3B &47.69 & 43.89 & 52.9 & 40.89 & 72.22 & 63.79 & 12.43 \\
    Qwen1.5 & 1.8B &46.55 & 46.71 & 37.88 & 39.43 & 60.3 & 61.42 & 33.59 \\
    Stablelm2 & 1.6B &45.25 & 38.95 & 43.34 & 36.78 & 64.56 & 70.45 & 17.44 \\
    \textbf{HARE} & 1.1B &40.17 & 35.74 & 38.4 & 42.08 & 59.27 & 57.46 & 8.04 \\
    H2o-danube & 1.8B &39.12 & 25.94 & 39.42 & 33.86 & 64.48 & 69.58 & 1.44 \\
    OpenELM & 1.1B &38.47 & 27.05 & 36.69 & 33.86 & 63.22 & 65.71 & 1.21 \\
    Csg-wukong & 1B &37.78 & 25.33 & 37.71 & 42.79 & 56.67 & 58.93 & 5.23 \\
    Cosmo & 1B & 36.59& 26.69 & 38.57 &38.15  & 55.49 & 55.13 & 5.53 \\
    TinyLlama & 1.1B &36.42 & 26.04 & 33.87 & 37.32 & 59.51 & 60.31 & 1.44 \\
    Gpt2xl &1.6B  &34.31 & 26.55 & 30.29 & 38.53 & 58.01 &51.39  & 1.06 \\
    \bottomrule[2pt]
  \end{tabular}
  \caption{\label{tab:results}
    Comparisons with the SOTA SLMs on Open LLM leaderboard. HARE performs favorably against existing SLMs in terms of average score.
  }
\end{table*}
\begin{table*}[ht]
  \centering
  \resizebox{0.9\textwidth}{!}{
  \begin{tabular}{c|cc|cc|cc|cc|cc|cc}
    \toprule[2pt]
    \multirow{2}{*}{\textbf{Model}} & \multicolumn{2}{c|}{\textbf{Gsm8k}}& \multicolumn{2}{c|}{\textbf{ARC}}& \multicolumn{2}{c|}{\textbf{Hellaswag}}& \multicolumn{2}{c|}{\textbf{MMLU}}& \multicolumn{2}{c|}{\textbf{Winogrande}}& \multicolumn{2}{c}{\textbf{TruthfulQA}} \\
     & $\Delta_{train}$ & $\Delta_{test}$& $\Delta_{train}$ & $\Delta_{test}$& $\Delta_{train}$ & $\Delta_{test}$& $\Delta_{train}$ & $\Delta_{test}$& $\Delta_{train}$ & $\Delta_{test}$& $\Delta_{train}$ & $\Delta_{test}$ \\
    \midrule[1pt]
    Phi1.5 &1.7 &0.62 &\textcolor{red}{21.8} &-0.15 &0.54 &- &- &\textcolor{blue}{-13.87} &\textcolor{red}{7.65} &- &\textcolor{red}{9.18} &- \\
    Qwen1.5 &\textcolor{red}{35.01} &\textcolor{red}{10.94} &0.51 &0.32 &0.94 &- &- &\textcolor{red}{1.85} &-0.13 &- &6.9 &- \\
    Stablelm2 &9.26 &0.28 &1.56 &\textcolor{red}{0.85} &\textcolor{red}{2.15} &- &- &-0.23 &-0.15 &- &3.85 &- \\
    \textbf{HARE} &7.63 &4.35 &\textcolor{blue}{-0.52} &\textcolor{blue}{-0.25} &1.31 &- &- &-2.61 &-0.16 &- &4.77 &- \\
    H2o-danube &0.29 &0.25 &0.01 &0 &\textcolor{blue}{0.03} &- &- &-0.5 &-0.02 &- &\textcolor{blue}{1.15} &- \\
    OpenELM &0.7 &0.9 &0.54 &0.25 &1.42 &- &- &-0.44 &-0.25 &- &9.14 &- \\
    Csg-wukong &11.27 &9.05 &0.37 &-0.22 &0.3 &- &- &-0.25 &-0.15 &- &4.33 &- \\
    Cosmo &\textcolor{blue}{0.24} &\textcolor{blue}{-0.71} &-0.11 &-0.03 &0.62 &- &- &0 &-0.22 &- &4.48 &- \\    
    TinyLlama &1.16 &1.03 &-0.29 &-0.05 &1.21 &- &- &0.11 &\textcolor{blue}{-0.32} &- &5.67 &- \\  
    Gpt2xl &0.74 &0.76 &0.55 &0.62 &0.49 &- &- &-0.33 &-0.25 &- &4.78 &- \\ 
    \bottomrule[2pt]
  \end{tabular}}
  \caption{\label{tab:data leak}
     Evaluation of benchmark data leakage using the method from \cite{xu2024benchmarking}. The greater the value of $\Delta$, the higher the probability of benchmark data leakage. \textcolor{red}{Red}: highest score, \textcolor{blue}{Blue}: lowest score. HARE maintains relatively low levels of $\Delta$ across comparison, indicating a lower probability of data leakage.
    }
      \vspace{-0.5cm}
\end{table*}

\subsection{Ablation studies}
Due to limited computing resources, we conduct ablation studies on a 0.25B model to efficiently validate the effectiveness of our proposed data construction method. We establish three experimental groups: (1) training on the $D_1$ dataset; (2) training on the combination of the $D_1$ and $D_2$ datasets; and (3) training on the integration of the $D_1$, $D_2$, and $D_3$ datasets. The three groups use the same parameter weights to initialize the model and the same hyperparameters during training, differing only in the composition of training data. The objective is to explore how different data combinations affect model performance.
Figure~\ref{fig:ablation_test} shows a gradual improvement in model performance with the addition of $D_2$ and $D_3$. This is because the addition of $D_2$ enhances the semantic diversity and consistency of data quality in the training dataset, while the inclusion of $D_3$ supplements the dataset with task-specific data for NLP tasks. Consequently, this improves the performance of our model on the benchmark datasets. The results of the ablation studies support using our final dataset, including $D_1$, $D_2$, and $D_3$, to train our HARE-1.1B model.

\subsection{Overall performance}
We compare our HARE-1.1B with 9 SLMs on the Open LLM Leaderboard and evaluate these models for benchmark data leakage using the method from \citep{xu2024benchmarking}. Appendix \ref{Data Leakage} presents the details of this method.
Table~\ref{tab:results} shows that HARE performs well on the Open LLM Leaderboard. 
Specifically, in terms of average score, our model outperforms models trained with web-scraped large-scale data, including \cite{zhang2024tinyllama, mehta2024openelm, singer2024h2odanube18b}, and models trained on synthetic data generated by LLMs lacking NLP-task data \cite{benallal2024cosmopedia}. These results demonstrate the effectiveness of our data construction method, which incorporates human priors in training data and enables our model to achieve favorable performance.
Although our model does not perform as well as the top 3 SLMs, Phi1.5, Qwen1.5, and Stablelm2, in terms of average score, we note that Table~\ref{tab:data leak} shows these models have significant $\Delta$ values on several benchmark train and test dataset evaluations, indicating a substantial risk of benchmark data leakage. In contrast, our model maintains relatively low $\Delta$ values across all benchmark evaluations.
The results from Table~\ref{tab:results} and Table~\ref{tab:data leak} effectively validate that our data construction method enhances model capabilities without introducing benchmark data leakage issues.

\section{Conclusion}
In this paper, we propose encoding human priors into data construction for training SLMs. Our data construction method ensures both semantic diversity and data quality consistency, while avoiding benchmark data leakage. We conducted extensive evaluations on various benchmarks to validate the effectiveness of the proposed method. The HARE-1.1B model, trained on the dataset built using this method, performs favorably against SOTA SLMs.

\section*{Limitations}
Our work explores the application of human priors in training SLMs, which entails certain limitations. Despite recognizing the importance of human priors, we lack a discussion on the quality of these priors. Inappropriately selected or biased human priors could lead to suboptimal or even misleading model outcomes, which will be explored and addressed in future work. Furthermore, due to constraints in computational resources, we were unable to fully explore the constraints between parameters of SLMs and human prior knowledge, leaving the qualitative relationship undetermined.

\section*{Acknowledgments}
Thanks to everyone who has contributed to this paper, including data collection and synthesis, model training, comparative experiments, and manuscript drafting. We are particularly thankful to Qun Zeng, who initially guided the direction of our work and provided meticulous guidance throughout. We also extend our appreciation to Shan Ma, Manjia Ding, and Daiyong Huang for their substantial support and help in our endeavors. The authors also wish to acknowledge the support of China Telecom Guizhou Branch for providing unwavering support.

\bibliography{custom}
\clearpage

\appendix

\section{Data Construction}
\subsection{Collected open-source data}
\label{Collected open-source data}
Table~\ref{tab:cleaned data} shows our collected open-source pre-training corpora. We clean these collected data to create $D_1$, with a sample  weight of 46.5\%.
\begin{table}[thbp]
  \centering
  \resizebox{\columnwidth}{!}{
  \begin{tabular}{c|c|c}
    \toprule[2pt]
    Source & Categories & Sample weights(\%) \\
    \midrule[1pt]
    RedPajamaC4 &C4 &11\%\\
    RedPajamaArxiv  &Arxiv &1.42\%\\
    Pubmed  &References &0.65\%\\
    S2orc  &Open Research Corpus &1.68\%\\
    PhiPapers  &Philosophy &0.06\%\\
    RedPajamaBook  &Book &1.73\%\\
    PG\_19   &Book &1.46\%\\
    RedPajamaStackExchange   &Net &1.10\%\\
    HackerNews   &News &0.10\%\\
    FreeLaw   &Law &0.46\%\\
    PileofLaw   &Law &0.34\%\\
    AMPS   &Khan Academy &0.96\%\\
    DM\_math &Math &0.5\%\\
    Orca\_math &Math &2.56\%\\
    OpenWebMath   &Math &3.88\%\\
    Fanfics   &Book &1.32\%\\
    RedPajamaWiki    &Wiki&7.78\%\\
    The-Stack    &Code &5.5\%\\
    StackOverFlow    &Code &4\%\\
    \bottomrule[2pt]
  \end{tabular}}
  \caption{\label{tab:cleaned data}
    \textbf{Open-source Pre-training Corpora Overview}: The open-source pre-training corpora are categorized.
  }
  \vspace{-0.5cm}
\end{table}

\subsection{Data synthesis using LLMs}
\label{Data synthesis using LLMs}
We sample seeds from the sources in $D_1$ shown in Table~\ref{tab:seed}.
Specific prompts are designed for different data sources, and Mixtral-8$\times$7B is used to generate synthetic data according to these prompts and seeds. We generate approximately 28 billion synthesized tokens. An example prompt used for generation is shown in this section. We combine our synthetic data with the cosmopedia dataset as $D_2$. The sample weights of our synthetic data and the cosmopedia dataset are 7.7\% and 5.93\%, respectively.
\begin{table}[thbp]
  \centering
  \resizebox{\columnwidth}{!}{
  \begin{tabular}{c|c|c}
    \toprule[2pt]
    Seed source & Categories & Synthetic tokens(B) \\
    \midrule[1pt]
    RedPajamaC4 & Web &4.5 \\
    RedPajamaWiki & Wiki &8.16 \\
    HackerNews & News &4.5 \\
    RedPajamaArxiv & Academic &4.62\\
    FanFics & Books &4.08 \\
    Pile-of-law & Law &2.16 \\
    The-Stack,StackOverFlow & Code &0.06 \\
    Open-Web-Math & Math &0.12 \\
    \bottomrule[2pt]
  \end{tabular}}
  \caption{\label{tab:seed}
  \textbf{Open-source Datasets for Data Synthesis}:
  Approximately 28B tokens for $D_2$ are synthesized based on $D_1$.
  }
\end{table}
\begin{center}
\begin{tcolorbox}
[colback=blue!5!white,colframe=blue!55!black,width=\columnwidth,title={Prompt for Law Data Generation},label={prompt_example_1}]
{   
Write a long and very detailed law course unit for a textbook, The course should be inspired from this text snippet:{\textbackslash}n"\{\}"{\textbackslash}n while trying to be: - Rigorous - you create challenging textbooks that cover the material in depth. {\textbackslash}n - Engaging - your textbooks have a narrative arc and engaging tone, like the writing of Michael Lewis. {\textbackslash}n - Applied - you use specific and practical examples. For example, if the topic is integration in calculus, include equations and proofs of the concept you're teaching. {\textbackslash}n As another example, if the topic is the history of the United States, include dates, names, and key events. Do not include a title or an introduction, simply write the content without headlines and introductory phrases. Focus purely on the subject itself without offering advice on teaching and instruction. The word count of the textbook needs to be greater than 1000 words. \\
}
\end{tcolorbox}
\end{center}

\subsection{Mixture of synthetic and open-source NLP task data}
\label{Mixture of Synthetic and Open-source NLP Task Data}
A subset of $D_2$ is sampled for synthesising NLP task data.
We select various NLP tasks, including Question-Answering (Q\&A), Multi-Choice Q\&A, Cloze, Summarization and more.
Multiple prompts are designed for each task, and Qwen-32B-Chat is used for data generation.
About 8 billion tokens are synthesized, and the synthetic NLP task data are referred to as "Restruct" in Table~\ref{tab:categories}.
An example prompt used for NLP task generation can be found in this secion.
We also collect open-source SFT data and make slight format modifications, converting it into various NLP task formats. This, combined with synthetic data, constitutes the final $D_3$ dataset.
The sample weights of the Restruct dataset and the open-source SFT dataset are 23.5\% and 16.37\%, respectively.
\begin{table}[thbp]
  \centering
  \resizebox{\columnwidth}{!}{
  \begin{tabular}{c|c|c}
    \toprule[2pt]
    Data & Source & Tokens(B) \\
    \midrule[1pt]
    Restruct & $D_2\textbackslash{Wiki}$, $D_2\textbackslash{Code}$, $D_2\textbackslash{Math}$, $D_2\textbackslash{Books}$, $D_2\textbackslash{Law}$ & 8 \\

    Open-source SFT & OpenHermes2.5, Auto-cot, Dolly and etc. & 10.97 \\

    \bottomrule[2pt]
  \end{tabular}}
  \caption{\label{tab:categories}
    \textbf{Mixture of Synthetic and Open-source NLP Task Data Overview}: Approximately 18.97B tokens of NLP task data are generated for $D_3$.
  }
\end{table}
\begin{center}
\begin{tcolorbox}
[colback=blue!5!white,colframe=blue!55!black,width=\columnwidth,title={Prompt for Multi-Choice Q\&A Data Generation},label={prompt_example_2}]
{
Please play the role of a data generator. Your task is to generate a data sample in the feild of \{topic\}.

Task Requirements:

 - You will be given a content about \{topic\}, you should read carefully and try to generate data with the content.

 - You will be given some examples, which you should learn and try to generate data in the form of the given examples.

 - Do NOT copy any of the examples given to you!!!

 - You should focus on the quality of the generated data samples not quantity.

Output Format:
Please output in the following format.
\{"question": "questions you generated", "options": ["option1", ..., "option4"], "answer": "gold option"\}

Content:
\{content\}

Examples:
\{examples\}
}
\end{tcolorbox}
\end{center}

\subsection{Data decontamination process}
\label{Data Decontamination Process}

All synthetic data will be rigorously decontaminated with benchmark datasets.
We first examined the statistical characteristics of the synthetic data, removing samples with excessively short or long context lengths.
In $D_2$, the text contextual length is restricted to approximately 2k tokens, while in $D_3$, the NLP task data is limited within 512 tokens.
Subsequently, we check the N-gram repetition rate between the generated data and benchmark datasets.
All samples in benchmark datasets are used for decontamination.
For long context length data in $D_2$, we set N in range of 10 to 15, while in $D_3$ is between 4 to 8.
Finally, We remove samples with a repetition rate exceeding 50\%, to ensure that all synthetic data are at no risk of data leakage.

\begin{table*}[!thbp]
  \centering
  \small
  \begin{tabular}{c|cccc|ccc}
    \toprule[2pt]
    \textbf{Task} & \textbf{Model} & $D^{'}_{train}$ & $D_{train}$ & $\Delta_{train}(\textbf{-})$ & $D^{'}_{test}$ & $D_{test}$ & $\Delta_{test}(\textbf{-})$ \\
    \midrule[1pt]
    \multirow{7}{*}{Gsm8k} & Phi1.5 & 15.59 & 17.29 & 1.7 & 14.86 & 15.48 & 0.62 \\
        & Qwen1.5 & 31.56 & 66.57 & 35.01 & 28.8 & 39.74 & 10.94  \\
        & Stablelm2 & 18.78 & 28.02 & 9.24 & 18.44 & 19.12 & 0.68 \\
        & \textbf{HARE} & 22.32 & 29.95 & 7.63 & 21.87 & 26.22 & 4.35 \\
        & H2o-danube & 0.34 & 0.63 & 0.29 & 0.34 & 0.59 & 0.25 \\
        & OpenELM & 10.23 & 10.93 & 0.7 & 10.34 & 11.24 & 0.9 \\
        & Csg-wukong & 11.32 & 22.59 & 11.27 & 11.04 & 20.09 & 9.05\\
        & Cosmo & 17.89 & 18.13 & 0.24 & 17.43 & 16.72 & -0.71\\
        & TinyLlama &8.99   &10.15 &1.16 & 9.24 & 10.27& 1.03 \\
        & Gpt2xl & 3.77 & 4.51 & 0.74 & 3.74 & 4.5 & 0.76 \\
    \midrule[1pt]
    \multirow{7}{*}{ARC-C} & Phi1.5 & 7.42 & 29.22 & 21.8 & 4.69 & 4.54 & -0.15 \\
        & Qwen1.5 & 1.3 & 1.81 & 0.51 & 1.3 & 1.62 & 0.32 \\
        & Stablelm2 & 1.89 & 3.45 & 1.56 & 2.2 & 3.05 & 0.85\\
        & \textbf{HARE} & 7.76 & 7.24 &-0.52 & 7.13 & 6.88 & -0.25 \\
        & H2o-danube & 0.04 & 0.05 &0.01  & 0.05 & 0.05 & 0 \\
        & OpenELM & 5.54 & 6.08 & 0.54 & 5.43 & 5.68 & 0.25 \\
        & Csg-wukong & 2.92 & 3.29 & 0.37 & 2.92 & 2.7 & -0.22 \\
        & Cosmo & 2.79 & 2.68 &-0.11 & 2.68 & 2.65 & -0.03 \\
        & TinyLlama &5.33   &5.04 &-0.29 & 5.05 & 5 & -0.05 \\
        & Gpt2xl &1.15 & 1.7 & 0.55 & 0.97 & 1.59 & 0.62 \\
    \midrule[1pt]
    \multirow{7}{*}{Hellaswag} & Phi1.5 &1.25  & 1.79 &0.54  & - & - & - \\
        & Qwen1.5 & 1.19  &2.13 &0.94 & - & - & - \\
        & Stablelm2 &1.36  &3.51 &2.15 & - & - & -\\
        & \textbf{HARE} &2.14  & 3.45 &1.31 & - & - & - \\
        & H2o-danube & 0 & 0.03 & 0.03 & - & - & - \\
        & OpenELM & 2.24 & 3.66 & 1.42 & - & - & - \\
        & Csg-wukong &1.33   & 1.63 &0.3 & - & - & - \\
        & Cosmo &1.45   & 2.07 & 0.62 & - & - & - \\
        & TinyLlama &2.05   &3.26 &1.21 & - & - & - \\
        & Gpt2xl &0.96 &1.45 & 0.49 & - & - & - \\
    \midrule[1pt]
    \multirow{7}{*}{Winogrande} & Phi1.5 & 11.13 & 18.78 & 7.65 & - & - & - \\
        & Qwen1.5 & 0.52 & 0.39 & -0.13 & - & - & - \\
        & Stablelm2 & 0.54 & 0.39 &-0.15 & - & - & -\\
        & \textbf{HARE} & 1.03 & 0.87 & -0.16  & - & - & - \\
        & H2o-danube & 0.02 & 0 & -0.02 & - & - & - \\
        & OpenELM & 1 & 0.75 & -0.25 & - & - & - \\
        & Csg-wukong & 0.43 & 0.28 & -0.15  & - & - & - \\
        & Cosmo &0.66   &0.44 &-0.22 & - & - & - \\
        & TinyLlama &0.96   &0.64 &-0.32 & - & - & - \\
        & Gpt2xl & 0.42 & 0.25 & -0.17 & - & - & - \\
    \midrule[1pt]
    \multirow{7}{*}{TruthfulQA} & Phi1.5 & 7.32 & 16.50 & 9.18 & - & - & - \\
        & Qwen1.5 & 4.85 & 11.75 & 6.9 & - & - & - \\
        & Stablelm2 & 3.4 & 7.25 &3.85 & - & - & -\\
        & \textbf{HARE} & 5.12 & 9.89 &4.77   & - & - & - \\
        & H2o-danube & 0.32 & 1.47 & 1.15 & - & - & - \\
        & OpenELM & 8.49 & 17.63 & 9.14 & - & - & - \\
        & Csg-wukong & 5.02 & 9.35 & 4.33  & - & - & - \\
        & Cosmo &5.43   &9.91 &4.48 & - & - & - \\
        & TinyLlama &6.81   &12.48 &5.67 & - & - & - \\
        & Gpt2xl &3.4 & 8.18 & 4.78 & - & - & - \\
    \midrule[1pt]
    \multirow{7}{*}{MMLU} & Phi1.5 & - & - & - & 39.47 & 25.67 & -13.87 \\
        & Qwen1.5 & -& - & - & 5.27 & 7.12 & 1.85 \\
        & Stablelm2 & - & - &- & 1.29 & 1.06 & -0.23\\
        & \textbf{HARE} & - & - &-   & 10.61 & 8 & -2.61 \\
        & H2o-danube & - & - & - & 2.17 & 1.67 & -0.5 \\
        & OpenELM & - & - & - & 1 & 0.56 & -0.44 \\
        & Csg-wukong & - & - & -  & 1.19 & 0.94 & -0.25 \\
        & Cosmo &-   &- &- & 0 & 0 & 0 \\
        & TinyLlama &-   &- &- & 0.95 & 1.06 & 0.11 \\
        & Gpt2xl & - & - & - & 0.33 & 0 & -0.33 \\
    \bottomrule[2pt]
  \end{tabular}
  \caption{\label{tab:data scores}
    Evaluation of benchmark data leakage using the method from \cite{xu2024benchmarking}. The greater the value of $\Delta$, the higher the probability of benchmark data leakage. HARE maintains relatively low levels of $\Delta$ across comparison, indicating a lower probability of data leakage.
    }
\end{table*}

\begin{table*}[!thbp]
  \centering
  \small
  \begin{tabular}{c|cccccccc}
    \toprule[2pt]
    \textbf{Model} & \textbf{Size} & \textbf{Average} & \textbf{MMLU} & \textbf{ARC-C} & \textbf{TruthfulQA} & \textbf{Winogrande} & \textbf{Hellaswag} & \textbf{Gsm8k} \\
    \midrule[1pt]
    Qwen1.5 & 1.8B & 43.99 & 45.87 & 38.74 & 40.62 & 59.67 & 60.02 & 19.03 \\
    \textbf{HARE} & 1.1B & 40.00 & 33.62 & 37.46 & 41.49 & 58.88 & 53.03 & 15.54 \\
    TinyLlama & 1.1B & 36.26 & 26.22 & 33.53 & 36.79 & 60.22 & 59.38 & 1.44 \\
    \bottomrule[2pt]
  \end{tabular}
  \caption{\label{tab:sft_results}
    \textbf{Results of Chat Models on Open LLM Leaderboard}: After SFT, HARE still maintains relatively competitive performance.
  }
\end{table*}

\begin{table*}[!thbp]
  \centering
  \resizebox{2.1\columnwidth}{!}{
  \begin{tabular}{c|c|c|c|c|c}
  \toprule[2pt]
    \multirow{2}{*}{\textbf{API Tasks}}
    &\multirow{2}{*}{\textbf{Descriptions of APIs}}
    &\multirow{2}{*}{\textbf{Training samples}}
    &\multirow{2}{*}{\textbf{Test samples}}
    &\multicolumn{2}{c}{\textbf{Accuracy(\%)}} \\
    & & & &\small{\textbf{HARE-1.1B-Tool}}&\small{Llama3-8B-RAG}\\
  \midrule[1pt]
    API-0 & Call the designated phone number & 690 & 50 & 91.7\% & 96\% \\
    API-1 & Call to the designated contact person & 732 & 50 & 100\% & 72\% \\
    API-2 & Search for specified content using a browser & 517 & 50 & 100\% & 32\% \\
    API-3 & Turn on the camera & 531 & 50 & 100\% & 90\% \\
    API-4 & Query weather & 514 & 50 & 100\% & 100\% \\
    API-5 & Send an email with given content to the designated recipient & 810 & 50 & 100\% & 100\% \\
    API-6 & Meaningless instructions & 469 & 50 & 95\% & 10\% \\
  \bottomrule[2pt]
  \end{tabular}}
  \caption{\label{tab:Android High-quality SFT data}\textbf{Android API Calling SFT Datasets and Evaluation}: Under seven API tasks, HARE-1.1B-Tool outperforms Llama3-8B-RAG in four tasks and exhibits a 26.1\% higher average accuracy.
  }
\end{table*}

\section{Data Leakage}
\label{Data Leakage}

We employ the method described in \citep{xu2024benchmarking} to evaluate benchmark data leakage. Specifically, we use GPT3.5-turbo to generate new datasets derived from the original benchmark datasets by modifying sentence structures or reordering answers. Subsequently, we assess the SLMs using accuracy scores for 5-grams with identical input to measure the performance differences between the original and new datasets. The evaluation scores are presented in Table~\ref{tab:data scores}.
$D^{'}$ and $D$ represent the performance of the models on synthetic and original datasets, respectively.
$\Delta$ measures the extent of memorization of the models on the original training or test datasets.
A higher $\Delta$ indicates deeper memorization of the training or test datasets, and a higher likelihood of data leakage.
When $\Delta$ is less than zero, it suggests that the model generalizes better on synthetic data. 
Since the comparison involves models of the same parameter scale, no normalization is applied to $\Delta$. 
Ideally, a model should perform similarly on both original and synthesized datasets, as reflected by $\Delta_{train}$ and $\Delta_{test}$. 
For a model that does not utilize train and test datasets, there will not be significant $\Delta$ due to changes in sentence descriptions or the order of options. 

In the evaluation on GSM8K, Qwen1.5 demonstrates significant $\Delta_{train}$ and $\Delta_{test}$ values, indicating a higher likelihood of data leakage.
During the evaluations of the ARC, Winogrande, and TruthfulQA datasets, Phi1.5 demonstrates a much greater $\Delta_{train}$ compared to similarly scaled SLMs, indicating an increased potential for data leakage.
Stablelm2 exhibits the largest $\Delta_{train}$ in the Hellaswag evaluation, indicating a higher possibility of data leakage.
In the evaluation of MMLU, we assess the zero-shot capabilities of models by taking the first non-empty character from model outputs as the response. Except for Qwen1.5 and TinyLlama-3T, models tend to prefer the phrasing and options of synthetic data, resulting in lower $\Delta$. Moreover, Phi1.5 and HARE demonstrate stronger zero-shot capabilities compared to other models.
Our model maintains relatively low levels of $\Delta$ across comparisons, indicating a lower probability of data leakage, while also demonstrating strong generalization capabilities across various datasets.
All datasets used in this evaluation will be made publicly available.

\section{Supervised Fine-tuning}
\label{Supervised Fine-tuning}
\subsection{Chat}
We fine-tune HARE-1.1B on a dataset comprising 0.25B tokens sourced from Dolly \citep{DatabricksBlog2023DollyV2}, MetaMathQA \citep{yu2023metamath}, UltraChat200k \citep{ding2023enhancing}, Auto-Cot \citep{zhang2023automatic}, and other collections.
All data are structured into a chat template featuring roles labeled as system, user, and assistant. 
Full fine-tuning is conducted on the model using 8$\times$A800 GPUs.
The learning rate is set to 1e-6, with a global batch size of 1024. 
Additionally, we assess the performance of the chat model on the open LLM leaderboard, with the results detailed in Table~\ref{tab:sft_results}.

\subsection{Android API calling}
We use the method proposed by \cite{chen2024octopus} to enable HARE-1.1B to have the capability to call Android APIs. 
The composition of the API Calling datasets is delineated in Table~\ref{tab:Android High-quality SFT data}. 
These datasets are synthesized utilizing the Llama3-8B \citep{touvron2023llama}, Kimi \cite{moonshot2023kimi}, and Mixtral-8$\times$7B \citep{jiang2024mixtral} models. 
Based on these datasets, we fine-tune HARE-1.1B to obtain the model HARE-1.1B-Tool, and successfully deploy its int4 quantized version on mobile devices. 
The inference speed reaches 8 tokens per second on a Xiaomi Redmi K40 phone.
Table~\ref{tab:Android High-quality SFT data} presents the comparative evaluation results of HARE-1.1B-Tool and Llama3-8B using RAG. The evaluation of Llama3-8B-RAG involves utilizing the RAG method to identify the most relevant function descriptions based on user queries. 
Subsequently, this language model uses these descriptions along with the user queries to generate the expected function call commands, thereby assessing its performance on test samples. 
The evaluation results for HARE-1.1B-Tool are obtained by fine-tuning on API Calling datasets and then conducting evaluations on test samples.

\end{document}